\documentclass[twocolumn,letterpaper]{IEEEAerospaceCLS}

\usepackage[utf8]{inputenc}
\usepackage{amsmath, amssymb, bm}
\usepackage{booktabs} 
\usepackage{graphicx}
\usepackage{float}

\usepackage[square,numbers]{natbib}

\usepackage[colorlinks=true, urlcolor=blue, linkcolor=black, citecolor=black]{hyperref}

\newcommand{\ignore}[1]{}
\newtheorem{prob}{\textbf{Problem}}
\begin{document}
\pagestyle{plain}
\title{Solving Geodesic Equations with Composite Bernstein Polynomials for Trajectory Planning}
\author{
    Nick Gorman\\
    Department of Mechanical Engineering\\
    University of Iowa\\
    Iowa City, IA 52240\\
    nicholas-gorman-1@uiowa.edu\\
    \and 
    Gage MacLin\\
    Department of Mechanical Engineering\\
    University of Iowa\\
    Iowa City, IA 52240\\
    gage-maclin@uiowa.edu\\
    \and 
    Maxwell Hammond\\
    Department of Mechanical Engineering\\
    University of Iowa\\
    Iowa City, IA 52240\\
    maxwell-hammond@uiowa.edu\\
    \and
    Venanzio Cichella\\
    Department of Mechanical Engineering\\
    University of Iowa\\
    Iowa City, IA 52240\\
    venanzio-cichella@uiowa.edu
    \thanks{\footnotesize This work was supported by NASA, ONR, and Amazon.}
    \thanks{\footnotesize 979-8-3315-7360-7/26/$\$31.00$ \copyright2026 IEEE}
}

\maketitle
\begin{abstract}
This work presents a trajectory planning method based on composite Bernstein polynomials, designed for autonomous systems navigating complex environments. The method is implemented in a symbolic optimization framework that enables continuous paths and precise control over trajectory shape. Trajectories are planned over a cost surface that encodes obstacles as continuous fields rather than discrete boundaries. Areas near obstacles are assigned higher costs, naturally encouraging the trajectory to maintain a safe distance while still allowing for efficient routing through constrained spaces. The use of composite Bernstein polynomials allows the trajectory to maintain continuity while enabling fine control over local curvature to meet geodesic constraints. The symbolic representation supports exact derivatives, improving the efficiency of the optimization process. The method is applicable to both two- and three-dimensional environments and is suitable for ground, aerial, underwater, and space system applications. In the context of spacecraft trajectory planning, for example, it enables the generation of continuous, dynamically feasible trajectories with high numerical efficiency, making it well suited for orbital maneuvers, rendezvous and proximity operations, operations in cluttered gravitational environments, and planetary exploration missions where on-board computational resources are limited. Demonstrations show that the approach can efficiently generate smooth, collision-free paths in scenarios with multiple obstacles, maintaining clearance without requiring extensive sampling or post-processing. The optimization is subject to several constraints: (1) a Gaussian surface implemented as an inequality constraint that ensures minimum clearance from obstacles; (2) geodesic equations are used such that the path follows the most efficient direction relative to the cost surface; (3) boundary constraints to enforce fixed start and end conditions. This approach can serve as a standalone planner or as an initializer for more complex motion planning problems. 
\end{abstract}
\tableofcontents
\section{Introduction}
Trajectory planning is a fundamental capability for autonomous systems. It enables these systems to move through environments safely and efficiently, with applications ranging from spacecraft orbital maneuvering and proximity operations \cite{barbee2010guidance}, self-driving vehicles \cite{selfdrivingCitation} and unmanned aerial vehicles (UAVs) \cite{aggarwal2020path} to robotic arms in manufacturing \cite{gasparetto2015path} and autonomous machinery in precision agriculture \cite{orchardPaper}. The heart of the challenge is to generate smooth paths that avoid obstacles, respect physical constraints, and lead the system from a starting point to a goal, all while also being efficient to compute and feasible to execute in real time.

Several broad classes of trajectory planning methods have been developed to address this challenge. Geometric approaches have been introduced for aerospace applications, using Hermite interpolation to enforce dynamic constraints at the boundaries \cite{patterson2023hermite}. Graph-based planners, such as A* \cite{hart1968formal} and Probabilistic Roadmaps (PRM) \cite{kavraki2002probabilistic}, discretize the space into a set of nodes and edges and search for a path using graph algorithms. These methods are fast and reliable but often produce jagged or suboptimal paths because they are limited by the resolution of the discretization \cite{lavalle2006planning}. Extensions to these methods include D* \cite{stentz1994optimal} for replanning in dynamic environments and Theta* \cite{Daniel_2010} for any-angle planning to reduce path artifacts. Hybrid approaches such as the work by Dolgov \cite{dolgov2010path} combine graph search with continuous optimization to improve solution quality in structured environments like road networks.

Sampling-based methods such as Rapidly-exploring Random Trees (RRT*) \cite{karaman2011sampling} avoid fixed grids by exploring the space through random sampling. These methods are more flexible and can scale to high-dimensional spaces, but they may struggle with convergence speed and often require post-processing to smooth out the resulting path. Real-time performance is not guaranteed, especially in cluttered or dynamic environments. Various extensions have been proposed to improve RRT* convergence, including informed sampling strategies \cite{gammell2014informed}, bidirectional variants \cite{kuffner2000rrt}, and kinodynamic extensions that incorporate system dynamics \cite{lavalle2001randomized}. For UAV applications specifically, methods like \cite{richter2016polynomial} use polynomial trajectory representations with RRT* to generate dynamically feasible paths. Despite these improvements, the fundamental reliance on sampling can limit solution quality in narrow passages or highly constrained spaces

Optimization-based planners take a different approach by formulating trajectory planning as a mathematical optimization problem. Examples include TrajOpt \cite{schulman2014motion}, which uses sequential convex optimization to avoid collisions, and minimum-jerk or minimum-snap planners used in quadrotor control \cite{mellinger2011minimum}. Differential flatness approaches \cite{van2012planning} exploit system structure to simplify trajectory generation for certain vehicle classes. CHOMP (Covariant Hamiltonian Optimization for Motion Planning) \cite{ratliff2009chomp} introduced functional gradient descent on trajectories represented as waypoints connected by cubic splines, while STOMP (Stochastic Trajectory Optimization for Motion Planning) \cite{kalakrishnan2011stomp} uses stochastic sampling to escape local minima. More recently, Model Predictive Path Integral (MPPI) methods \cite{williams2017information} have shown promise for real-time replanning under uncertainty. These methods can produce very smooth and dynamically feasible paths, but they require careful formulation and usually depend on a good initial guess to converge successfully \cite{kelly2017introduction, kielas2022bernstein}.

This work introduces a method based on geodesic equations, which define the "straightest possible paths" within a curved space. Geodesics are fundamental in fields like general relativity \cite{o1983semi}, where they describe how particles move through curved spacetime, and they also provide a powerful tool for trajectory planning. By defining a cost surface that encodes obstacles, assigning higher cost values near obstacles and lower values in free space, the geodesic equations naturally guide the trajectory to avoid high-cost regions while maintaining smoothness \cite{do1992riemannian}.  

Geodesic-based path planning has been explored in several contexts. In computer graphics and surface processing, geodesics on triangulated meshes are computed using methods such as the fast marching method \cite{kimmel1998computing} and heat method \cite{crane2013geodesics}. For robotics, configuration space metrics have been designed to encode kinematic and dynamic constraints \cite{jaillet2010path}, enabling geodesic paths that respect system limitations. Riemannian Motion Policies \cite{cheng2019riemannian} use learned metrics for manipulation planning. However, these approaches typically require discretized representations or iterative numerical integration of geodesic equations, limiting their applicability for real-time trajectory generation in continuous spaces with complex obstacle configurations.


However, solving geodesic equations directly is difficult due to their nonlinear structure and the complexity of arbitrary cost surfaces. To address this, we propose a formulation using composite Bernstein polynomials \cite{maclin2024optimal}, which represent the trajectory as a piecewise-smooth polynomial curve. By employing composite Bernstein polynomials of arbitrary order, we can tune the optimality of the approach, adjusting the approximation capability of the trajectory representation as needed. This contrasts with approaches such as Pythagorean hodographs \cite{farouki1990pythagorean,choe2015trajectory}, which typically rely on polynomials of fixed order and therefore offer less flexibility. The method is implemented within the CasADi symbolic optimization framework \cite{andersson2019casadi}, allowing for exact computation of derivatives and efficient enforcement of the geodesic equations. The result is a flexible method for generating smooth, obstacle-avoiding trajectories in N-dimensional spaces which can be solved rapidly and, e.g., used as a feasible initial guess for an optimal motion planner in tight, crowded environments. It combines the geometric insight of configuration-space planners with the precision and tunability of optimal control methods \cite{lavalle2006planning, betts2010practical}. Compared to A*, RRT*, and TrajOpt, our approach offers continuous representations, guaranteed constraints satisfaction enforced geometrically, and higher numerical efficiency through symbolic computation. 

This work presents a novel formulation that combines continuous Gaussian cost fields with direct geodesic computation through symbolic optimization using composite Bernstein polynomials. Unlike discrete representations or sampling-based approaches, our method represents obstacles as continuous fields and solves for trajectories that are geodesic with respect to this surface, enabling precise constraint satisfaction and efficient computation.

The paper is organized as follows. In Section 2, we introduce the optimal control problem and the discretized nonlinear programming problem via composite Bernstein polynomials. Then in Section 3, we describe the construction of a Gaussian cost surface and the computation of a geodesic curve that navigates this surface. In Section 4, we solve a 2D and 3D geodesic- and geodesic-like nonlinear program via the composite Bernstein direct method, and then investigate the benefits of using these solutions to initialize more complex optimal control problems. Lastly, in Section 5, concluding remarks are made.

\section{Direct Approximation of Optimal Control Problems}
\subsection{Optimal Control Problem}
We seek to solve the optimal trajectory planning problem, i.e., find the optimal state $x^*(t)$ and control $u^*(t)$ based on some specified cost function, while considering vehicle and mission constraints. Thus, we formally define the general optimal control problem (OCP) as follows:
\begin{prob}\label{prob:OCP}
	\begin{equation}
	\begin{split}  
	& \min_{x(t),u(t),{{t}_f}}  J(x(t),u(t),{{t}_f}) = \\  &  E(x(0),x 
 (t_f),t_f)+ \int_0^{t_f} F(x(t),u(t))dt \, 
	\end{split} 
	\end{equation}
subject to 
	\begin{align}
	& x'(t) = f(x(t),u(t))\, , \quad \forall t \in[0,t_f], \label{eq:dynamicconstraint} \\
	& e(x(0),x(t_f),{{t}_f}) = 0 \, , \label{eq:equalityconstraint} \\
	& h(x(t),u(t)) \leq 0 \, , \quad \forall t\in [0,t_f] \label{eq:inequalityconstraint} \, ,
	\end{align}
\end{prob}

where $J$ represents the Bolza-type cost function, $t_f$ is the final mission time, $E$ is the initial and end point cost, and $F$ is the running cost, \eqref{eq:dynamicconstraint} governs the dynamics of the system, boundary conditions are imposed in \eqref{eq:equalityconstraint}, and inequality constraints are enforced in \eqref{eq:inequalityconstraint}. For simplicity, $x(t)\in\mathbb{R}$ and $u(t)\in\mathbb{R}$, however Problem \ref{prob:OCP} readily extends to higher dimensions.

To solve this problem numerically, we must discretize the OCP into a nonlinear programming (NLP) problem which can then be solved using off-the-shelf software.

\subsection{Composite Bernstein Polynomials}
The discretization scheme used in this paper is based on the Bernstein polynomial basis \cite{farouki2012bernstein}. An $N$th order Bernstein polynomial $x_N:[t_0,t_f]\to\mathbb{R}$ can be expressed as
\begin{equation}
    x_N(t)=\sum^N_{j=0}\bar{x}_{j,N}b_{j,N}(t), \quad t\in[t_0,t_f]
\end{equation}
where $\bar{x}_{j,N}\in\mathbb{R},\,j=0,...,N$ are the Bernstein coefficients which we hereafter refer to as control points, and where $b_{j,N}(t)$ is the Bernstein basis function
\begin{equation}
    b_{j,N}(t)=\binom{N}{j}\frac{(t-t_0)^j(t_f-t)^{N-j}}{(t_f-t_0)^N},
\end{equation}
where $\binom{N}{j}$ are the binomial coefficients. The control points of $x_N(t)$ can be organized into a vector as $\bar{\bm{x}}_N=[\bar{x}_{0,N},...,\bar{x}_{N,N}]\in\mathbb{R}^{N+1}$.

The Bernstein basis was selected due to its many beneficial properties for trajectory planning, in particular, geometrically enforced boundary conditions and inequality constraints, as well as constraint enforcement along the entire polynomial as opposed to only at the control points \cite{cichella2020optimal}. 

We now introduce a composite Bernstein polynomial as a series of Bernstein polynomials connected together sharing terminal control points, referred to as knots. Composite Bernstein polynomials $x_M:[t_0,t_f]\to\mathbb{R}$ consist of $K$ Bernstein polynomials defined over the subintervals $[t_{k-1},t_k],\, k=1,...,K$, where $t_0<t_1<...<t_K$ and $t_K=t_f$, and take the form of
\begin{equation}
x_M(t)=
    \begin{cases}
        x_N^{[1]}(t)=\sum^N_{j=0}\bar x_{j,N}^{[1]}b_{j,N}^{[1]}(t), \quad t\in[t_0,t_1] \\
        x_N^{[2]}(t)=\sum^N_{j=0}\bar x_{j,N}^{[2]}b_{j,N}^{[2]}(t), \quad t\in[t_1,t_2] \\
        \vdots \\
        x_N^{[K]}(t)=\sum^N_{j=0}\bar x_{j,N}^{[K]}b_{j,N}^{[K]}(t). \, t\in[t_{K-1},t_K] \\
    \end{cases}
\end{equation}
These approximants inherit the beneficial geometric properties of Bernstein polynomials, while converging at a much faster rate \cite{maclin2024optimal}.

\subsection{Integration-based Discretization of OCPs}
Bernstein approximation of optimal control problems is typically done via direct collocation of the $N+1$ control points onto the approximated optimal solution. As the non-boundary control points do not lie on the resulting polynomial, convergence can be slow. With composite Bernstein approximation, convergence is much faster when compared to single Bernstein approximation, with a convergence rate of $\mathcal{O}\left(\frac{1}{K^2N}\right)$ as opposed to $\mathcal{O}\left(\frac{1}{N}\right)$. However, a composite Bernstein polynomial has $K(N+1)$ control points, and to take full advantage of the faster convergence rate the number of total control points would significantly increase, limiting computational benefits. Alternatives to this control point collocation approach include knots-only collocation \cite{hammond2025solutions}, wherein we only optimize over the knots and then compute the remaining control points via integration, and network-approximated OCPs \cite{maclin2025real}, where a network is trained on optimal data.

The control points of the integral of a composite Bernstein polynomial can be computed algebraically as a matrix-vector product
\begin{equation}\label{eq:integral}
    \bar{\bm x}_M=\bar{\bm{x}}'_M\bm{\Gamma}_M+c,
\end{equation}
where $\bar{\bm{x}}_M=[\bar{\bm{x}}_N^{[1]},\bar{\bm{x}}_N^{[2]},...,\bar{\bm{x}}_N^{[K]}]\in\mathbb{R}^{K(N+1)}$ is a vector of the control points of the composite Bernstein polynomial, with bold letters denoting vectors or matrices, $c\in\mathbb{R}$ is an integration constant, and $\bm \Gamma_M \in\mathbb{R}^{K(N+1)\times K(N+2)}$ is the composite Bernstein integration matrix
\begin{equation}\label{Integral}
    \bm{\Gamma}_M = 
    \begin{bmatrix}
        \bm{\gamma}_N^{[1]} & \bm{\phi}_{1,N} & \hdots & \hdots & \bm{\phi}_{1,N} \\
        \bm{0} & \bm{\gamma}_N^{[2]} & \bm{\phi}_{2,N} & \hdots & \bm{\phi}_{2,N} \\
        \vdots & \ddots & \ddots & \ddots & \vdots \\
        \vdots & \ddots & \ddots & \ddots & \bm{\phi}_{K-1,N} \\
        \bm{0} & \hdots & \hdots & \bm{0} & \bm{\gamma}_N^{[K]}
    \end{bmatrix},
\end{equation}
where $\bm{\gamma}_N^{[k]}\in\mathbb{R}^{(N+1)\times(N+2)}$ is the Bernstein integration matrix for the $k$th segment
\begin{equation}\label{eq:BernsteinIntegration}
    \bm{\gamma}_N^{[k]}=
    \begin{bmatrix}
        0 & \frac{t_k-t_{k-1}}{N+1} & \hdots & \frac{t_k-t_{k-1}}{N+1} \\
        \vdots & \ddots & \ddots & \vdots \\
        0 & \hdots & 0 & \frac{t_k-t_{k-1}}{N+1} \\
    \end{bmatrix},
\end{equation}
and $\bm{\phi}_{k,N}\in\mathbb{R}^{(N+1)\times (N+2)}$ enforces continuity between segments and is defined as
\begin{equation}\label{eq:phi}
    \bm{\phi}_{k,N}=\frac{t_{k}-t_{k-1}}{N+1}
    \begin{bmatrix}
        1 & 1 & \hdots & 1 \\
        \vdots & \ddots & \ddots & \vdots \\
        1 & \hdots & 1 & 1
    \end{bmatrix}.
\end{equation}

Noticing that the integration property of Bernstein polynomials increases the polynomial order by one, we can represent any composite Bernstein polynomial in terms of its zeroth order $M$th derivative and its initial conditions. This manipulation reduces the number of necessary optimization variables from $K(N+1)$ to $K+M$ (as $N=0$), significantly reducing the computation time for high $K$. These unknown variables are denoted as
\begin{equation}
    \bm{\theta}_M=\left[\bar{\bm{x}}_M^{(M)},c_0,c_1,...,c_{M-1}\right] \in\mathbb{R}^{K+M}.
\end{equation}
Then the derivatives of $\bar{\bm{x}}_M$ can be obtained in the form $\bm\theta_m$ where $m=M,...,0$ by
\begin{equation}\label{eq:zeta}
\begin{split}
    \bm\theta_{M-1}=\bm\theta_M\bm\zeta_M, \\
    \vdots \\
    \bm\theta_0=\bm\theta_{1}\bm\zeta_{1},
\end{split}
\end{equation}
such that, e.g., the control points $\bar{\bm{x}}_M$ and $\bar{\bm{x}}_M^{(1)}$ are contained in $\bm\theta_0$ and $\bm\theta_1$ respectively, and where $\bm\zeta_m\in\mathbb{R}^{(K(m+1)+M)\times(K(m+2)+M)}$ is a modified integration matrix that maintains the structure of the initial conditions and allows for the above operations. The construction of which can be found in Appendix \ref{appendix:CBP}.

As we are only collocating the knots of $\bar{\bm{x}}_M^{(M)}$ (i.e., the knots of the $M$th derivative of the composite Bernstein polynomial $\bm{x}_M(t)$) onto the NLP, we must be able to obtain the knots of $\bar{\bm{x}}_M^{(m)}$ from $\bm\theta_M$ for $m=0,...,M$. To do this, we can use the following operation
\begin{equation}
    \bar{\bm{X}}_M^{(m)}=\bm\theta_M\bm{T}_m,
\end{equation}
where $\bar{\bm{X}}_M^{(m)}\in\mathbb{R}^{K+1}$ are the knots of $\bar{\bm{x}}_M^{(m)}$ for $m=0,...,M$ and $\bm{T}_m\in\mathbb{R}^{(K+M)\times(K+1)}$ is a cumulative integral matrix
\begin{equation}
    \bm{T}_m=\bm{P}_m\bm\zeta_m...\bm\zeta_M,
\end{equation}
and $\bm{P}_m=\{p_{i,j}\}\in\mathbb{R}^{(K(m+1)+M)\times (K+1)}$ is a selector matrix with
\begin{equation}
\begin{split}
    p_{i,j}=
    \begin{cases}
        1 & i=j=0 \\
        1 & i=k(m+1)-1,\, j=k \\
        0 & \text{otherwise}
    \end{cases}
    \\k=1,...,K.
\end{split}
\end{equation}
Now we can formally define the NLP as:
\begin{prob}\label{prob:NLP}
\begin{equation} \label{eq:costfunc}
	\begin{split}  
	& \min_{\bm\theta_{M,x},\bm\theta_{M,u},{{t}_K},t_0}  J\left(\bm\theta_{M,x},\bm\theta_{M,u},{{t}_K},t_0\right) = \\  &  E\left(\bar{\bm{X}}_{0,M}^{(0)},\bar{\bm{X}}_{K,M}^{(0)}
 ,t_K\right)+ \frac{t_K-t_0}{K+1}\sum^K_{k=0}F\left(\bar{\bm{X}}_{k,M}^{(0)},\bar{\bm{U}}_{k,M}^{(0)}\right)
	\end{split} 
\end{equation}
subject to 
	\begin{align}
	& ||\bar{\bm{X}}_M^{(1)}-\bm{f}(\bar{\bm{X}}_M^{(0)},\bar{\bm{U}}_M^{(0)})||\le\bm\delta_M \\
	& \bm{e}(\bar{\bm{X}}_{0,M}^{(0)},\bar{\bm{X}}_{K,M}^{(0)}
 ,t_K) = \bm{0} \, ,  \\
	& \bm{h}(\bar{\bm{X}}_M^{(0)},\bar{\bm{U}}_M^{(0)}) \leq \bm{0}  \, ,
	\end{align}
\end{prob}
where $\bm\theta_{M,x}$ and $\bm\theta_{M,u}$ are the optimization variables for the states and controls respectively,
$\bar{\bm{X}}_{j,M}^{(0)}$ represents the $j$th knot of $\bar{\bm{X}}_M^{(0)}$. Lastly, $\delta_M$ is a relaxation bound equal to a small positive number that depends on $K$ and converges uniformly to 0, i.e., $\text{lim}_{K\to\infty}\delta_M=0.$

%
%

\section{Motion Planning Using Gaussian Surfaces and Geodesics}
The problem of finding an object-avoiding trajectory in crowded environments can be challenging to solve using traditional optimization-based approaches. Alternatively, we propose that these crowded environments be represented as Gaussian surfaces, i.e., each obstacle is represented as a peak on the surface. Then, a geodesic trajectory can be computed which innately avoids these peaks, and thus the real obstacles.
\subsection{Encoding Obstacle 
Fields as Gaussian Surfaces}
For the 2D case, we define the cost surface as:
\begin{equation}
f(u, v) = A \sum_{i=1}^N \exp\left(-\kappa\frac{(u - x_{o,i})^2 + (v - y_{o,i})^2}{2 r_i^2}\right),
\end{equation}
where \( (x_{o,i}, y_{o,i}) \) are obstacle centers, \( r_i \) is the miniumum distance from the object, \( A \) is the amplitude, \( N \) is the number of obstacles, and $\kappa$ is a tunable parameter that adjusts the sharpness of each peak. In 3D, the cost surface extends to include a \( z \)-coordinate:
\begin{multline}
f(u, v, w) = \\A \sum_{i=1}^N \exp\left(-\kappa\frac{(u - x_{o,i})^2 + (v - y_{o,i})^2+(w - z_{o,i})^2}{2 r_i^2}\right)
\end{multline}
In 2D, this surface meets the criteria for a "Monge patch". A Monge patch is a surface described as a height field over a 2D Cartesian plane. In 2D, the surface is defined over the \( u \)-\( v \) plane as
\[
\mathbf{x}(u, v) = \begin{bmatrix} u \\ v \\ f(u, v) \end{bmatrix},
\]
where \( f(u, v) \) gives the height of the surface above each point \( (u, v) \). This concept generalizes naturally to 3D, where the hypersurface is embedded in four-dimensional space and defined over the \( u \)-\( v \)-\( w \) volume as
\[
\mathbf{x}(u, v, w) = \begin{bmatrix} u \\ v \\ w \\ f(u, v, w) \end{bmatrix}.
\]

Now that we can construct a Monge patch with peaks and valleys that represent some obstacle field, we introduce the geodesic equation to navigate this surface.

\subsection{The Geodesic Equation}

In this setting, the quantity
\begin{equation}
g = 1 + \sum_{i=1}^D \left( \frac{\partial f}{\partial x_i} \right)^2,
\end{equation}
represents the squared norm of the normal vector to the surface or hypersurface, with \( x=(u,v) \) when $D=2$ or \( (u,v,w) \) when $D=3$.

Christoffel symbols , denoted \( \Gamma^k_{ij} \), describe how the coordinate basis vectors of the tangent space of a manifold changes from point to point. The symbol \( \Gamma^k_{ij} \) describes the similarity between tangent basis vectors \(x_i\) and \(x_k\) as they change in the \(x_j\) coordinate. They encode how curved the space is and are essential for expressing derivatives of vector fields in curved spaces. Specifically, Christoffel symbols appear in the geodesic equation, whose solutions are curves that can locally minimize distance. 

In the context of a Monge patch with cost function \( f \), the Christoffel symbols can be written as:
\begin{equation}
\Gamma^k_{ij} = \frac{1}{g} \, \frac{\partial^2 f}{\partial x_i \partial x_j} \, \frac{\partial f}{\partial x_k}, \quad i, j, k = 1, \dots, D,
\end{equation}
where the Einstein summation convention is used (i.e., repeated indices are implicitly summed over).

The geodesic equation on a manifold with coordinates \( x_1, \dots, x_D \) is given by:
\begin{align}
\frac{d^2 x^k}{dt^2} + \Gamma^k_{ij} \frac{dx^i}{dt} \frac{dx^j}{dt} &= 0, \quad \text{for } k = 1, \dots, D
\end{align}
or, explicitly, in 2D:
\begin{align}
\begin{split}
x_1'' + \Gamma^1_{11} (x_1')^2 
+ 2\Gamma^1_{12} x_1' x_2' 
+ \Gamma^1_{22} (x_2')^2 
= 0,
\end{split} \\
\begin{split}
x_2'' + \Gamma^2_{11} (x_1')^2 
+ 2\Gamma^2_{12} x_1' x_2' 
+ \Gamma^2_{22} (x_2')^2 
= 0,
\end{split}
\end{align}
and in 3D:
\begin{align}
\begin{split}
x_1'' + \Gamma^1_{11} (x_1')^2 
+ 2\Gamma^1_{12} x_1' x_2' 
+ 2\Gamma^1_{13} x_1' x_3' 
+ \Gamma^1_{22} (x_2')^2 \\
+ 2\Gamma^1_{23} x_2' x_3' 
+ \Gamma^1_{33} (x_3')^2 
= 0,
\end{split} \\
\begin{split}
x_2'' + \Gamma^2_{11} (x_1')^2 
+ 2\Gamma^2_{12} x_1' x_2' 
+ 2\Gamma^2_{13} x_1' x_3' 
+ \Gamma^2_{22} (x_2')^2 \\
+ 2\Gamma^2_{23} x_2' x_3' 
+ \Gamma^2_{33} (x_3')^2 
= 0,
\end{split} \\
\begin{split}
x_3'' + \Gamma^3_{11} (x_1')^2 
+ 2\Gamma^3_{12} x_1' x_2' 
+ 2\Gamma^3_{13} x_1' x_3' 
+ \Gamma^3_{22} (x_2')^2 \\
+ 2\Gamma^3_{23} x_2' x_3' 
+ \Gamma^3_{33} (x_3')^2 
= 0,
\end{split}
\end{align}
where primes denote time derivatives. 

\begin{figure}[htbp]
    \centering
    \includegraphics[width=\linewidth,trim=0cm 0cm 0cm 1.0cm,clip]{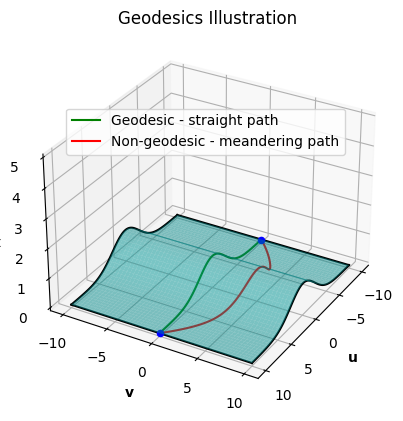}
    \caption{Geodesic vs. Non-Geodesic.}
    \label{fig:geodesicIllustration}
    
\end{figure}

This system of second-order differential equations describes the motion of a particle along a curve that follows the most natural or straightest possible path on the surface, determined entirely by the geometry of the space. It does so by eliminating any tangential acceleration—ensuring that the particle's velocity changes only in the normal direction, not along the surface itself. As a result, the path has no intrinsic turning or swiveling at any point. The difference between a geodesic curve and a non-geodesic curve traversing a surface is illustrated in Figure \ref{fig:geodesicIllustration}.

\begin{figure}[htbp]
    \centering
    \includegraphics[width=\linewidth,trim=0cm 0cm 0cm 1.0cm,clip]{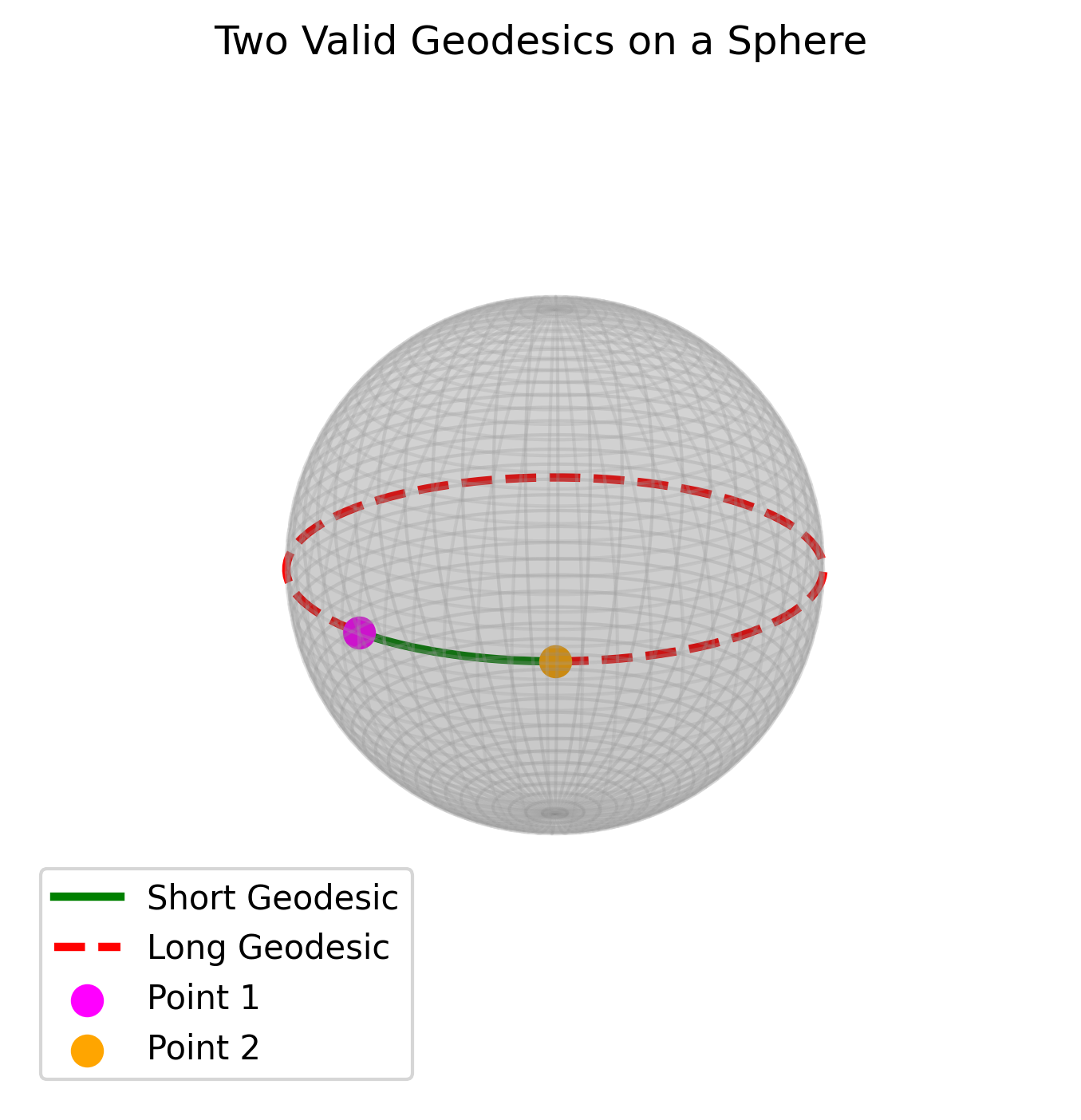}
    \caption{Two Valid Geodesics.}
    \label{fig:greatArcs}
\end{figure}

While such geodesic curves often minimize distance locally, they do not always correspond to globally shortest paths, especially in curved or complex spaces. For example, on the surface of a sphere, the shortest path between two points lies along a great circle. However, between any two non-antipodal points, there are two great circle arcs connecting them—one shorter and one longer—both of which satisfy the geodesic equation, though only the shorter one represents the true minimal-distance path. These great arcs are shown in Figure \ref{fig:greatArcs}.

To study the efficacy of modeling object fields as a Gaussian surface navigated by a geodesic trajectory, we introduce the geodesic equations and the cost surface as constraints into the previously defined NLP, and parameterize the trajectory using composite Bernstein polynomials.

\section{Numerical Results}
All results in this section were solved on a Dell Precision 7670 laptop with a 12th Gen Intel(R) Core(TM) i7-12850HX processor clocked at 2.10 GHz and 32GB of installed RAM. No steps were taken for hardware acceleration. 

\subsection{Numerical Results: Geodesic Solutions}

\subsubsection{Solving the 2D Geodesic Equations with Composite Bernstein Polynomials}
The geodesic trajectory planning problem was implemented using the composite Bernstein direct method and solved in CasADi using symbolic variables. Obstacle centers and radii were encoded in a Gaussian cost surface \( f \), and the geodesic equations were created symbolically via automatic differentiation in CasADi. The trajectory was parameterized with \( M = 3 \), \( K = 45 \) and \( D=2 \), yielding
\begin{equation*}
\bm\theta_{M,x}\in\mathbb{R}^{D\times(K+M)}
\end{equation*}

decision variables. $M$ was chosen s.t. $\bar{\bm{x}}_M^{(M)}$ is one order greater than the second-order system being solved. Geodesic equations and boundary conditions are entered as equality constraints while inequality constraints enforced the obstacle avoidance criterion \( f(x_1,x_2) \leq \rho \), with \( \rho \approx A e^{-\kappa/2} \). The selection of parameter \(\rho\) is described in Appendix \ref{appendix:rho}.

The 2D OCP representing a geodesic trajectory navigating a Gaussian cost surface is presented as
\begin{prob}\label{prob:OCP2D}
\begin{equation}
\begin{aligned}
\min_{x_1,x_2} \quad & \int^{t_f}_{t_0}((x'_1)^2+(x'_2)^2)dt, & \\
\text{subject to} \quad & x_1'' + \Gamma^1_{11} (x_1')^2  + 2\Gamma^1_{12} x_1' x_2' + \Gamma^1_{22} (x_2')^2 = 0, \\
& x_2'' + \Gamma^2_{11} (x_1')^2 + 2\Gamma^2_{12} x_1' x_2' + \Gamma^2_{22} (x_2')^2 = 0, \\
& x_1(t_0)=x_{1,0}, \quad x_1(t_f)=x_{1,f}, \\
& x_2(t_0)=x_{2,0}, \quad x_2(t_f)=x_{2,f}, \\
& f(x_1,x_2) \leq \rho. \\
\end{aligned}
\end{equation}
\end{prob}

%
This formulation includes the geodesic constraints, the Gaussian cost surface, and specified initial and final conditions. The cost function for the above problem was selected to minimize the accumulated square of the arc length along the trajectory. This construction ensures that the trajectory will be as short as possible while still honoring the geodesic and Gaussian constraints. To ensure that the initial guess to the problems satisfies the boundary conditions, a straight line guess can be used. However, as we are optimizing over the control points of the $M$th derivative, the guess needs to be constructed using $\bm\theta_M$. The control points of the initial guess $\bm\theta_M$ were all set to zero, and the initial velocity was simply approximated by the total distance between boundary conditions divided by the mission time. Due to this initialization, the resulting $\bar{\bm{x}}_M^{(0)}$ is a straight line that satisfies the desired boundary conditions. Recall that $\bar{\bm{x}}_M^{(0)}$ can be extracted from $\bm\theta_0$, which is obtained via the successive integration of $\bm\theta_M$.

\begin{figure}[htbp]
    \centering
    \includegraphics[width=\linewidth,trim=0cm 0cm 0cm 0.75cm,clip]{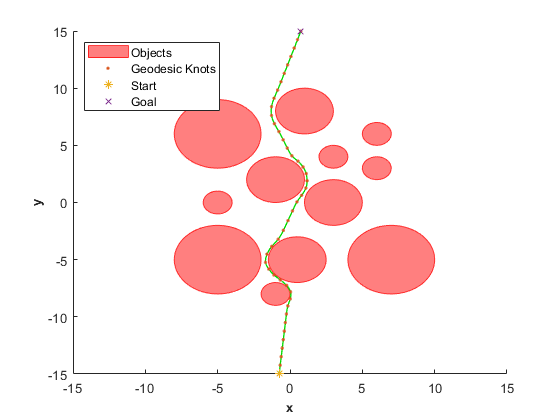}
    \caption{An example geodesic-constrained trajectory through a 12-obstacle environment.}
    \label{fig:traj2d}
\end{figure}

\begin{figure}[htbp]
    \centering
    \includegraphics[width=\linewidth,trim=0cm 0cm 0cm 0.75cm,clip]{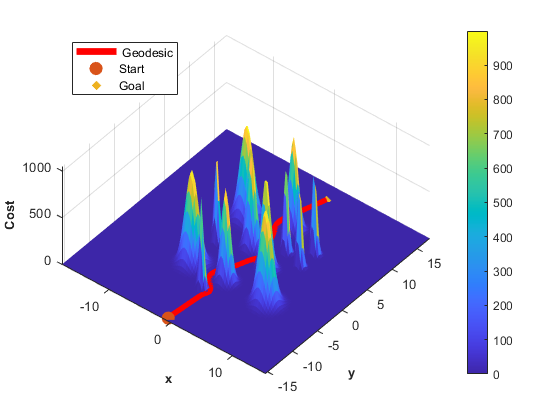}
    \caption{An example geodesic-constrained trajectory shown on top of the associated obstacles' cost surface.}
    \label{fig:surf}
\end{figure}

Problem \ref{prob:OCP2D} was discretized into the form of Problem \ref{prob:NLP} and solved using CasADi’s IPOPT \cite{andersson2019casadi} with the MA57 solver \cite{duff2004ma57}, with decision variable \( \bm\theta_M \) and a straight line initial guess, with the resulting trajectory shown in Figure \ref{fig:traj2d}. This trajectory is able to successfully navigate the object field, attaining the goal position. This same trajectory can be seen navigating the Gaussian cost surface in Figure \ref{fig:surf}.

CasADi’s symbolic framework was used to compute the Christoffel symbols analytically and to construct the full system of geodesic equations. This symbolic representation enables the generation of exact Jacobians and Hessians, which are supplied to IPOPT to improve convergence and numerical stability during optimization. Using symbolic differentiation improves both performance and accuracy. Unlike numerical gradients, which can suffer from round-off error or require additional function evaluations, CasADi generates efficient computational graphs for evaluating derivatives exactly at any point in the optimization. This improves the rate of convergence, reduces the number of iterations, and ensures consistent constraint enforcement across all knot points.

To evaluate the method's reliability and runtime performance, 250 trajectory optimization problems were solved with randomly selecting pairs of points on a circle that fully encloses all twelve obstacles in the environment as boundary conditions for varying $K$. Each trial involved solving the discretized version of Problem \ref{prob:OCP2D}. The amplitude of $f$ was set at 1000 and the variance adjustment factor was set to 10. The Gaussian amplitude being set so high enables the solver to make subtle tweaks in the solution and place the knots at exactly the right place to attain the curvature needed for a solution. The cost surface and its relation to obstacle clearance, as detailed in Appendix~\ref{appendix:3DGauss}, were used to determine the value of \(\rho\) for the cost inequality constraint. By enforcing \(f \leq \rho\) at the trajectory knots, the path avoids obstacles.

Performance was evaluated at seven resolution levels by varying the number of segments \(K \in \{9, 15, 21, 27, 33, 39, 45\}\). As expected, higher-resolution trajectories incurred greater computational cost with the benefit of higher accuracy and a lower failure rate. The results of this experiment can be found in Table \ref{tab:geo2d}. The \textit{Solve (s)} refers to the average time required to solve the full geodesic-constrained NLP. This time excludes symbolic expression generation and is the solve time only. The \textit{Infeasible} column contains how many computations failed because the solver reported the problem as infeasible. It should be noted that many of these trajectories are still object avoiding, but computation failed due to not satisfying the geodesic constraints. From a warmstarting perspective, these infeasible, though obstacle-avoiding trajectories are still viable initial guesses.


\begin{table}[h]
\centering
\caption{Optimization Performance from Solving the 2D Geodesic-Constrained NLP 250 Times for Each $K$}
\label{tab:geo2d}
\scriptsize
\setlength{\tabcolsep}{2pt}
\begin{tabular}{c|c|ccc}
\toprule
\textbf{K} & \textbf{Solve (s)}  & \textbf{Infeasible} \\
\midrule
9  & 0.33   & 99 \\
15 & 0.65   & 86  \\
21 & 1.25   & 74  \\
27 & 1.52   & 71  \\
33 & 4.26   & 65  \\
39 & 5.08   & 61  \\
45 & 6.65   & 71  \\
\bottomrule
\end{tabular}
\end{table}

\subsubsection{Solving the 3D Geodesic Equations Using Composite Bernstein Polynomials} To determine the efficacy of this approach for more complicated problems, we formulate a 3D variant of Problem \ref{prob:OCP2D} that uses the 3D geodesic equations and the 3D cost surface as constraints: 
\begin{prob}\label{prob:OCP3D}
\begin{equation}
\begin{aligned}
\min_{x_1,x_2,x_3} \quad & \int^{t_f}_{t_0}((x'_1)^2+(x'_2)^2+(x'_3)^2)dt, & \\
\text{subject to} \quad 
& x_1'' + \Gamma^1_{11} (x_1')^2 
+ 2\Gamma^1_{12} x_1' x_2' 
+ 2\Gamma^1_{13} x_1' x_3'\\&\quad
+ \Gamma^1_{22} (x_2')^2 
+ 2\Gamma^1_{23} x_2' x_3' 
+ \Gamma^1_{33} (x_3')^2 
= 0, \\
& x_2'' + \Gamma^2_{11} (x_1')^2 
+ 2\Gamma^2_{12} x_1' x_2' 
+ 2\Gamma^2_{13} x_1' x_3'  \\&\quad
+ \Gamma^2_{22} (x_2')^2
+ 2\Gamma^2_{23} x_2' x_3' 
+ \Gamma^2_{33} (x_3')^2 
= 0, \\
& x_3'' + \Gamma^3_{11} (x_1')^2 
+ 2\Gamma^3_{12} x_1' x_2' 
+ 2\Gamma^3_{13} x_1' x_3' \\&\quad
+ \Gamma^3_{22} (x_2')^2 
+ 2\Gamma^3_{23} x_2' x_3' 
+ \Gamma^3_{33} (x_3')^2 
= 0. \\
& x_1(t_0)=x_{1,0}, \quad x_1(t_f)=x_{1,f}, \\
& x_2(t_0)=x_{2,0}, \quad x_2(t_f)=x_{2,f}, \\
& x_3(t_0)=x_{3,0}, \quad x_3(t_f)=x_{3,f}, \\
& f(x_1,x_2,x_3) \leq \rho. \\
\end{aligned}
\end{equation}
\end{prob}

\begin{figure}[htbp]
    \centering
    \includegraphics[width=\linewidth,trim=0cm 0cm 0cm 0.75cm,clip]{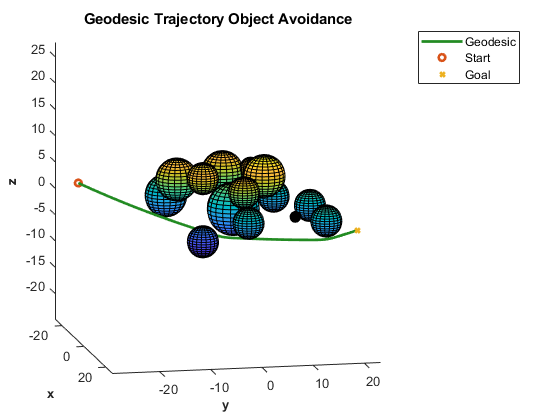}
    \caption{An example trajectory shown avoiding 15 objects.}
    \label{fig:3Dgeodesic}
\end{figure}

Problem \ref{prob:OCP3D} was discretized in the form of Problem \ref{prob:NLP} and solved for \( M = 3 \), \( K = 45 \) and \( D=3 \), with $\rho=1000$ and $\kappa=10$. An example of this is shown in Figure \ref{fig:3Dgeodesic}.

As previously, we solve the discretized version of Problem \ref{prob:OCP3D} 250 times for each \(K \in \{9, 15, 21, 27, 33, 39, 45\}\), with randomly selected pairs of boundary conditions and 15 obstacles. The results can be seen in Table \ref{tab:geo3d}, with markedly fewer infeasible solutions than the 2D case. Notably, however, the solver returns a large amount of infeasible trajectories for both cases, particularly at low $K$. This is because for low discretization, the control points of adjacent polynomials may be incompatible with the local geodesic constraints i.e. one polynomial starts to affect the other. At higher K, more granular control allows for geodesic constraints to be satisfied. This infeasibility indicates that a geodesic trajectory was not found for the specific random boundary conditions.

\begin{table}[h]
\centering
\caption{Optimization Performance from Solving the 3D Geodesic-Constrained NLP 250 Times for Each $K$}
\label{tab:geo3d}
\scriptsize
\setlength{\tabcolsep}{2pt}
\begin{tabular}{c|c|ccc}
\toprule
\textbf{K} & \textbf{Solve (s)}  & \textbf{Infeasible} \\
\midrule
9  & 0.74  & 45 \\
15 & 1.73  & 35 \\
21 & 3.96  & 29 \\
27 & 4.06  & 18 \\
33 & 5.09  & 12 \\
39 & 6.18  & 6  \\
45 & 9.19  & 8  \\
\bottomrule
\end{tabular}
\end{table}

\subsection{Numerical Results: Geodesic-Like Solutions}

To investigate the potential benefits of using a geodesic-constrained trajectory as an initial guess to solve a more complex optimal motion planning problem, i.e., "warmstarting", we construct an additional OCP for comparison. This problem is similar to Problems \ref{prob:OCP2D} and \ref{prob:OCP3D}, in that the obstacle field is represented as a soft constraint via the Gaussian cost surface. However, the hard geodesic constraint is removed from this formulation, resulting in a trajectory that is geodesic-like, in that the arc length is still minimized. 
%
%

\subsubsection{Generating Trajectories Without Geodesic Constraints}
Problem \ref{prob:OCP2D} was discretized and solved (without the hard geodesic constraint) using a straight line guess with the same parameters and $K$ values as in the previous section. Two random points on a circle completely encompassing the object field were selected as boundary conditions. This process was repeated 250 times for each value of $K$, with results summarized in Table \ref{tab:nongeo2d}. The above steps were followed for trajectories in 3D as well, instead solving Problem \ref{prob:OCP3D} with random boundary conditions selected along the surface of the encompassing sphere, with results provided in Table \ref{tab:nongeo3d}. From these two tables, it is immediately apparent that without the geodesic constraint, the solve time is reduced by over an order of magnitude. This is as expected as the OCP is massively simplified when compared to the geodesic-constrained version.

\begin{table}[h]
\centering
\caption{Optimization Performance from Solving the 2D Geodesic-Like NLP 250 Times for Each $K$}
\label{tab:nongeo2d}
\scriptsize
\setlength{\tabcolsep}{2pt}
\begin{tabular}{c|c|ccc}
\toprule
\textbf{K} & \textbf{Solve (s)}& \textbf{Infeasible} \\
\midrule
9  & 0.04  & 0  \\
15 & 0.04  & 0  \\
21 & 0.06  & 0  \\
27 & 0.06  & 0  \\
33 & 0.12  & 0  \\
39 & 0.13  & 0  \\
45 & 0.16  & 0  \\
\bottomrule
\end{tabular}
\end{table}


\begin{table}[h]
\centering
\caption{Optimization Performance from Solving the 3D Geodesic-Like NLP 250 Times for Each $K$}
\label{tab:nongeo3d}
\scriptsize
\setlength{\tabcolsep}{2pt}
\begin{tabular}{c|c|cccc}
\toprule
\textbf{K} & \textbf{Solve (s)}  & \textbf{Infeasible} \\
\midrule
9  & 0.04 & 0 \\
15 & 0.06 & 0 \\
21 & 0.06 & 0 \\
27 & 0.07 & 0 \\
33 & 0.09 & 0 \\
39 & 0.12 & 0 \\
45 & 0.14 & 0 \\
\bottomrule
\end{tabular}
\end{table}

\subsection{Numerical Results: Warmstarting an OCP}
Now that we can compute geodesic-constrained and geodesic-like trajectories, we want to investigate the benefits of using these candidate trajectories as initial guesses to an OCP that considers vehicles dynamics and hard collision avoidance constraints. To provide a fair baseline, a straight-line guess will also be used in the initialization for comparison.


\begin{prob}\label{prob:OCPdubins}
\begin{equation}
\begin{aligned}
\min_{x_1,x_2} \quad & \int^{t_f}_{t_0}((x'_1)^2+(x'_2)^2)dt, & \\
\text{subject to} \quad 
& x_1(t_0)=x_{1,0}, \quad x_1(t_f)=x_{1,f}, \\
& x_2(t_0)=x_{2,0}, \quad x_2(t_f)=x_{2,f}, \\
& \sqrt{(x_1-x^i_{1,\text{obs}})^2+(x_2-x^i_{2,\text{obs}})^2}\ge r_{\text{sep}},\\
& \forall i=1,...,N_{\text{obs}}\\
& \sqrt{(x_1')^2+(x_2')^2}\le V,
\end{aligned}
\end{equation}
\end{prob}



where $x^i_{1,\text{obs}},x^i_{2,\text{obs}}$ are the center position of each obstacle, $r_\text{sep}$ is the obstacle radius, $N_\text{obs}$ is the number of obstacles, and $V$ is some maximum velocity. Problem \ref{prob:OCPdubins} was discretized and solved with randomized boundary conditions, 750 times each for various values of $K$, with 250 times for each initialization process: geodesic-constrained, geodesic-like, and straight-line guess. Table \ref{tab:solver_compare_bc2} displays the results of these simulations, with the average solve time including the computation time to solve Problem \ref{prob:OCPdubins}, as well as the time required to generate the initial guess. As evident from Table \ref{tab:solver_compare_bc2}, the solution found using the straight line guess is significantly faster at low $K$, though the computation time rapidly increases as $K$ increases. Using the geodesic-constrained trajectories to warmstart Problem \ref{prob:OCPdubins} begins to reduce computation time at $K=39$ when compared to the straight-line guess. However, the geodesic-like are shown to be even better candidates for warmstarting, being faster than both alternatives starting at $K=27$.

\begin{table}[h]
\centering
\caption{Average Solve Times Across 250 Random Trajectories Traversing the Obstacle Field for Each $K$, with Various Initial Guesses: Geodesic-Constrained, Geodesic-Like, and Straight Line}
\label{tab:solver_compare_bc2}
\scriptsize
\setlength{\tabcolsep}{2pt}
\begin{tabular}{c|ccc}
\toprule
\textbf{K} & \textbf{Geo Time (s)} & \textbf{Geo-Like Time (s)} & \textbf{Line Time (s)} \\
\midrule
9  & 1.94 & 1.58 & 0.04 \\
15 & 2.27 & 1.57 & 0.36 \\
21 & 2.64 & 1.64 & 1.27 \\
27 & 3.33 & 1.69 & 2.82 \\
33 & 5.83 & 1.85 & 5.00 \\
39 & 5.91 & 1.98 & 8.16 \\
45 & 8.80 & 2.28 & 13.31 \\
\bottomrule
\end{tabular}
\end{table}
\section{Conclusions}
This work demonstrates a method for trajectory planning using composite Bernstein polynomials with soft obstacle avoidance constraints. We consider (i) geodesic-constrained trajectories which naturally follow the straightest path along a surface, and (ii) geodesic-like trajectories, which omit the explicit geodesic constraint but use a length-minimizing cost function. For collision avoidance, the obstacle field is encoded as a Gaussian surface. With the introduction of an inequality constraint limiting the surface height, the paths are constrained to navigate the Gaussian surface without collision. All expressions are generated symbolically to improve numerical accuracy and solver speed. These trajectories can be used directly or to warmstart more complex motion-planning problems. We find that using these geodesic-like trajectories  as a warmstart can significantly improve computation time, particularly at high orders of approximation.

Tests in a 2D environment with twelve obstacles show that the method reliably produces paths that avoid all obstacles. The IPOPT solver reaches an optimal solution without constraint violations, confirming both accuracy and stability. 3D results also suggest that this geodesic approach to path planning can be extended to higher dimensions without major changes. Because the formulation is dimension-agnostic, the same structure can be applied to more complex, higher dimensional planning problems. 
Future work will focus on improving runtime, implementing different cost surfaces instead of Gaussians (a 3D Witch of Agnessi is a good example), handling dynamic obstacles, and adding real-time implementation to make the method more useful for real-world robots and autonomous systems. 

\appendices{}
\section{Encoding Distance into Gaussian Variance In 2D Using Gaussian Curvature} 
\label{appendix:rho}
For a single obstacle in 2D, the cost surface is
\begin{equation}\label{eq:simpleCost}
    f(u, v) = A \exp\left(-\kappa\frac{u^2 + v^2}{\sigma}\right),
\end{equation}  
where the variance is related to the obstacle radius by:
\begin{equation}
\frac{\sigma}{\kappa} = 2 r^2.
\end{equation}
and \(\kappa\) is a variable used to adjust the sharpness of the object Gaussians to minimize the "mixing" of objects close to each other.

The Gaussian curvature for a Monge patch \( x(u, v) = [u, v, f(u, v)]^\top \) is:
\begin{equation}
K_G = \frac{f_{uu} f_{vv} - f_{uv}^2}{(1 + f_u^2 + f_v^2)^2},
\end{equation}
as derived in classical differential geometry texts \cite{pressley2010elementary}.
For the cost surface in \eqref{eq:simpleCost}, this is:
\begin{equation}
\begin{split}
&K_G = \\&\, \frac{4 a^{2} \kappa^{2} \left(- 4 \kappa^{2} u^{2} v^{2} + \left(2 \kappa u^{2} - \sigma\right) \left(2 \kappa v^{2} - \sigma\right)\right) e^{\frac{2 \kappa \left(u^{2} + v^{2}\right)}{\sigma}}}{\left(4 a^{2} \kappa^{2} u^{2} + 4 a^{2} \kappa^{2} v^{2} + \sigma^{2} e^{\frac{2 \kappa \left(u^{2} + v^{2}\right)}{\sigma}}\right)^{2}}.
\end{split}
\end{equation}
Setting $K_G = 0$ yields:
\begin{equation}
u^2 + v^2 = \frac{\sigma}{2 \kappa}.
\end{equation}
Because \( r^2 = u^2 + v^2 \), this yields the relation of variance to distance:
\begin{equation}
\frac{\sigma}{\kappa} = 2r^2,
\end{equation}
which is the radius where curvature changes sign. At this distance, the cost is:
\begin{equation}
\rho = A e^{-\frac{\kappa}{2}},
\end{equation}
Setting \( \rho \approx A e^{-\frac{\kappa}{2}} \)  ensures trajectories stay at least distance \( r \) from obstacles, with a safety margin. By enforcing \( f \leq \rho \) at \( K \) collocation points, the trajectory navigates around obstacles, maintaining a safe distance.

\section{Encoding Distance into Gaussian Variance in 3D Using Gauss-Kronecker Curvature} \label{appendix:3DGauss}

For a single obstacle in 3D, the cost surface is
\begin{equation}\label{eq:simpleCost3D}
    f(u, v, w) = A \exp\left(-\kappa\frac{u^2 + v^2 + w^2}{\sigma}\right),
\end{equation}
where the variance is related to the obstacle radius by:
\begin{equation}
\frac{\sigma}{\kappa} = 2 r^2,
\end{equation}
and \(\kappa\) adjusts the sharpness of the obstacle Gaussian.

The Gauss–Kronecker curvature of a hypersurface in \(\mathbb{R}^4\) defined by a Monge patch 
\(\mathbf{x}(u, v, w) = [u, v, w, f(u,v,w)]^\top\)
is
\begin{equation}
K_{GK} = \frac{\det(\text{Hess}\,f)}{(1 + f_u^2 + f_v^2 + f_w^2)^{5/2}},
\end{equation}
where \(\text{Hess}\,f\) is the Hessian matrix of \(f(u, v, w)\).

For the cost surface in \eqref{eq:simpleCost3D}, the first derivatives are:
\begin{align}
f_u &= -\frac{2 A \kappa u}{\sigma} 
       \exp\left(-\kappa \frac{u^2 + v^2 + w^2}{\sigma}\right), \\
f_v &= -\frac{2 A \kappa v}{\sigma} 
       \exp\left(-\kappa \frac{u^2 + v^2 + w^2}{\sigma}\right), \\
f_w &= -\frac{2 A \kappa w}{\sigma} 
       \exp\left(-\kappa \frac{u^2 + v^2 + w^2}{\sigma}\right).
\end{align}
The second derivatives (Hessian terms) are:
\begin{align}
f_{uu} &= \frac{2 A \kappa}{\sigma} 
          \left(\frac{2 \kappa u^2}{\sigma} - 1\right)
          \exp\left(-\kappa \frac{u^2 + v^2 + w^2}{\sigma}\right), \\
f_{vv} &= \frac{2 A \kappa}{\sigma} 
          \left(\frac{2 \kappa v^2}{\sigma} - 1\right)
          \exp\left(-\kappa \frac{u^2 + v^2 + w^2}{\sigma}\right), \\
f_{ww} &= \frac{2 A \kappa}{\sigma} 
          \left(\frac{2 \kappa w^2}{\sigma} - 1\right)
          \exp\left(-\kappa \frac{u^2 + v^2 + w^2}{\sigma}\right).
\end{align}

Setting \(K_{GK} = 0\) and solving for \(\sigma\) yields:
\begin{equation}
\sigma = 2 \kappa (u^2 + v^2 + w^2).
\end{equation}
Defining \(r^2 = u^2 + v^2 + w^2\) gives:
\begin{equation}
\frac{\sigma}{\kappa} = 2 r^2,
\end{equation}
which defines a spherical surface of radius
\begin{equation}
r = \sqrt{\frac{\sigma}{2 \kappa}}.
\end{equation}
At this distance, the cost evaluates to
\begin{equation}
\rho = A e^{-\frac{\kappa}{2}}.
\end{equation}

Setting \(\rho \approx A e^{-\frac{\kappa}{2}}\) ensures that trajectories remain at least distance \(r\) from obstacles, with a safety margin. By enforcing 
\(f \leq \rho\) at collocation points, the trajectory navigates around 3D obstacles while maintaining a safe clearance.

\section{Composite Bernstein Integration Matrix}\label{appendix:CBP}
The modified integration matrix presented in \eqref{eq:zeta} is constructed as
\begin{equation}
    \bm\zeta_M=
    \begin{bmatrix}
        \bm\Gamma_M & \bm0 \\
        \bm\Psi_M & \bm I_M
    \end{bmatrix}\in\mathbb{R}^{(K(N+1)+M)\times(K(N+2)+M)},
\end{equation}
where $\bm\Gamma_M$ is the composite Bernstein integration matrix
\begin{equation}\label{Integral2}
    \bm{\Gamma}_M = 
    \begin{bmatrix}
        \bm{\gamma}_N^{[1]} & \bm{\phi}_{1,N} & \hdots & \hdots & \bm{\phi}_{1,N} \\
        \bm{0} & \bm{\gamma}_N^{[2]} & \bm{\phi}_{2,N} & \hdots & \bm{\phi}_{2,N} \\
        \vdots & \ddots & \ddots & \ddots & \vdots \\
        \vdots & \ddots & \ddots & \ddots & \bm{\phi}_{K-1,N} \\
        \bm{0} & \hdots & \hdots & \bm{0} & \bm{\gamma}_N^{[K]}
    \end{bmatrix},
\end{equation}
with $\bm\gamma_N^{[k]}$ and $\bm\phi_{k,N}$ defined in \eqref{eq:BernsteinIntegration} and \eqref{eq:phi} respectively. To properly add the integration constants to the appropriate control points, the submatrix $\bm\Psi_M$ is defined as
\begin{equation}
    \bm\Phi_M=\{\psi_{i,j}\}\in\mathbb{R}^{M\times K(N+2)},
\end{equation}
where
\begin{equation}
    \psi_{i,j}=\begin{cases}
        1 & i=N+1 \\
        0 & \text{otherwise}
    \end{cases}.
\end{equation}
Lastly, $\bm I_M$ is the identity matrix in $\mathbb{R}^{M\times M}$ which preserves the initial conditions. This results in a modified integration matrix $\bm\zeta_M$ that maintains continuity at knots and respects the structure of the integration constants, allowing for operations such as $\bm\theta_0=\bm\theta_1\bm\zeta_1$.

\providecommand{\newblock}{}
\bibliographystyle{IEEEtran}
\bibliography{refs}

\thebiography
\begin{biographywithpic}
{Nick Gorman}{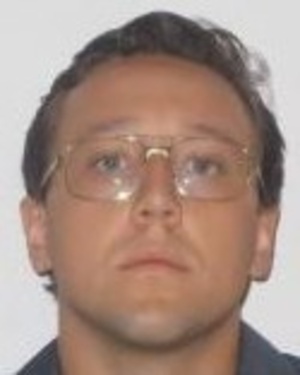}
received his B.S. and MS degrees in electrical engineering from the University of Iowa. He is currently a Project Engineer in the Datalinks RF Group at Collins Aerospace. He previously held a position and Lutron Electronics on their RF Team designing wireless dimming and lighting controls. He has also spent three years conducting research in Professor Cichella's Cooperative Autonomous System laboratory at the University of Iowa.
\end{biographywithpic}

\begin{biographywithpic}
{Gage MacLin}{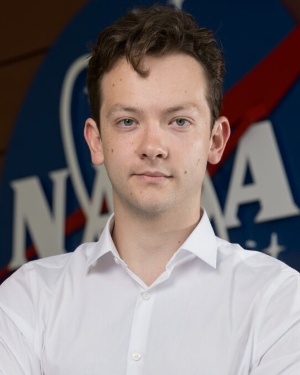}
received the B.S. and M.S. degrees in mechanical engineering from the University of Iowa, Iowa City, Iowa, USA in 2023 and 2024 respectively, and is pursuing a Ph.D. at the same institution. His research interests include numerical methods for optimal control, learning-based trajectory generation and multi-agent path planning.
\end{biographywithpic}

\begin{biographywithpic}
{Maxwell Hammond}{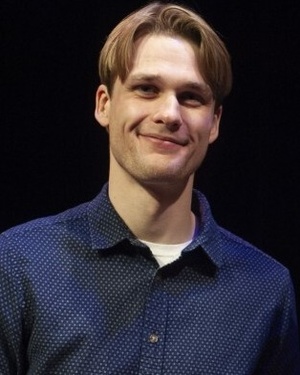}
received the B.S., M.S., and Ph.D. degrees in mechanical engineering from the University of Iowa, Iowa City, Iowa, USA in 2019, 2020, and 2025 respectively. His research interests include optimal path planning, PDE-based control, control of soft robotics, and control of continuum systems.
\end{biographywithpic}

\begin{biographywithpic}
{Venanzio Cichella}{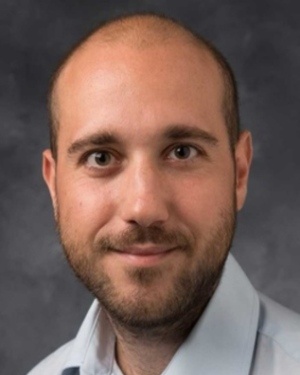}
received the B.S. and M.S.
degrees in automation engineering from the
University of Bologna, Bologna, Italy, in 2007
and 2011, respectively and the Ph.D. degree
in mechanical science and engineering from
the University of Illinois at Urbana-Champaign,
Champaign, IL, USA, majoring in optimal control, cooperative control, and autonomous systems in 2018.
Since 2018, he has been with the Mechanical Engineering Department, University of Iowa,
Iowa City, IA, USA, where he is currently an Associate Professor of
robotics and autonomous systems, and the Director of the Cooperative
Autonomous Systems (CAS) Lab. His research interests include cooperative control of autonomous robots, optimal control, nonlinear systems,
robust and adaptive control, and human-centered robotic design.
\end{biographywithpic}

\end{document}